\documentclass[10pt, journal, compsoc]{IEEEtran}
\ifCLASSOPTIONcompsoc
\usepackage[nocompress]{cite}
\else
\usepackage{cite}
\fi
 
\ifCLASSINFOpdf
    \usepackage[pdftex]{graphicx}
    \DeclareGraphicsExtensions{.pdf,.jpeg,.png}
\else
   \usepackage[dvips]{graphicx}
  \DeclareGraphicsExtensions{.eps}
\fi

\ifCLASSOPTIONcompsoc
  \usepackage[caption=false,font=footnotesize,labelfont=sf,textfont=sf]{subfig}
\else
  \usepackage[caption=false,font=footnotesize]{subfig}
\fi

\usepackage{etex}
\usepackage{booktabs} 
\usepackage{url}
\usepackage[ruled,linesnumbered]{algorithm2e}
\usepackage{algpseudocode}
\usepackage{float}
\usepackage{array}
\usepackage{adjustbox}
\usepackage{multirow}
\usepackage{diagbox}
\usepackage{amsmath}
\usepackage{amsfonts}
\usepackage{subfig}
\usepackage{tcolorbox} 
\usepackage{cleveref}
\usepackage{balance}

\crefname{figure}{Figure}{Figures}
\graphicspath{{figs/}}

\usepackage{tikz}
\usepackage{tikz,fullpage}
\usepackage{tkz-berge}
\usetikzlibrary{arrows,shapes,decorations,automata,backgrounds,petri}

\makeatletter
\renewcommand\paragraph{\@startsection{paragraph}{4}{\z@}%
	{1.5ex plus .2ex minus .3ex}%
	{-0em}%
	{\normalsize\bf}}
\makeatother
\makeatletter
\newcommand{\removelatexerror}{\let\@latex@error\@gobble}
\makeatother

\title{Informative Path Planning for Location Fingerprint Collection}

\author{Yongyong~Wei,~Cristian~Frincu,~Rong~Zheng,~\IEEEmembership{Senior~Member,~IEEE}
		\IEEEcompsocitemizethanks{\IEEEcompsocthanksitem Yongyong Wei, Cristian Frincu and Rong Zheng are with the Department of Computing and Software, McMaster University, Hamilton, ON, Canada.\protect\\
			E-mails:  \{{\it weiy49, frincuc, rzheng}\}@mcmaster.ca
			}
	}
	
\begin{document}

\IEEEtitleabstractindextext{%
\begin{abstract}
Fingerprint-based indoor localization methods are promising due to the high availability of deployed access points and compatibility with commercial-off-the-shelf user devices. However, to train regression models for localization, an extensive site survey is required to collect fingerprint data from the target areas. In this paper, we consider the problem of informative path planning  (IPP) to find the optimal walk for site survey subject to a budget constraint. IPP for location fingerprint collection is related to the well-known orienteering problem (OP) but is more challenging due to edge-based non-additive rewards and revisits.  Given the NP-hardness of IPP, we propose two heuristic approaches: a Greedy algorithm and a genetic algorithm. We show through experimental data collected from two indoor environments with different characteristics that the two algorithms have low computation complexity, can generally achieve higher utility and lower localization errors compared to the extension of two state-of-the-art approaches to OP. 

\end{abstract}

}

\maketitle
\IEEEdisplaynontitleabstractindextext

\IEEEraisesectionheading{\section{Introduction}}
\label{sect:intro}
 
\IEEEPARstart{I}{ndoor} localization using location dependent fingerprints and already existing infrastructure such as cellular, or WiFi networks is an attractive solution due to low deployment cost. In general, a fingerprint-based solution works in two stages: training and operation. In the training stage, a comprehensive site survey is conducted to record the fingerprints at targeted locations and a machine learning model (typically a regression model) is trained afterwards. The site survey process needs to be done repeatedly to account for any changes in the infrastructure and environment. In the operation stage, when a user submits a location query with her currently observed fingerprints, a location server computes and returns her estimated location.  

Lack of efficient means to collect extensive fingerprints is preventing the technology from becoming common place. To lower the barrier of site surveys, researchers have developed methods that leverage crowd sourcing~\cite{yang2012locating} or take advantage of other phone sensors to annotate fingerprint data with location labels when users walk along self-selected paths~\cite{li2017turf}. While those approaches address the question of {\it how} data is collected, they do not consider {\it where} data is collected. As a result, fingerprints are either obtained opportunistically, which likely leaves many areas unsurveyed, or exhaustively following some regular grid pattern which is  time consuming for large buildings. 

In this paper, we investigate Informative Path Planning (IPP) for location fingerprint site survey (short as {\it site survey} in the rest of the paper) in indoor environments. Combined with the path-based fingerprint data collection approach in~\cite{li2017turf}, we believe that we are one step closer to realizing practical fingerprint based indoor positioning systems.

IPP aims to generate a walk to collect data that is deemed to contain the most information (and thus most ``informative'') under cost constraints. In IPP, both the utility of data and costs are application specific. In site surveys, it is natural to associate costs with travelling time. An obvious candidate for utility is the location error of a trained model from the collected data. Unfortunately, such a utility measure is intractable analytically, sensitive to the regression model used, and dependent on both the locations where data were collected, and the actual measurements from these locations. The last aspect makes localization errors an ill fit for path planning that needs to be conducted {\it offline} and ahead of the site survey. 

Following the work in~\cite{li2017turf}, a Gaussian Process (GP)~\cite{rasmussen2006gaussian} (also known as a Kriging model) is used to model location fingerprints as samples from a random process in space. GPs have been shown to be effective at modeling wireless signals~\cite{hahnel2006gaussian}\cite{schwaighofer2004gpps}\cite{li2017turf}, terrain~\cite{vasudevan2009gaussian} and water temperature~\cite{grbic2013stream} among many others. As a metric for utility, we choose mutual information (MI) as a surrogate for localization errors. MI quantifies how much one set of random variables tells us about another set.  

IPP on a graph with vertex rewards \footnote{In this paper, we use the terms ``utility'' and ``reward'' interchangeably.} given a time budget constraint is also called the orienteering problem (OP)~\cite{vansteenwegen2011orienteering}. OP is known to be NP-hard. What distinguishes IPP for location fingerprint collected from OP is three-fold. First, under MI, the utility function is not additive. Second, the utility is a function of the edges along the selected path not of the vertices. This is because fingerprints are either collected at fixed sampling intervals (e.g., for magnetic sensing) or roughly periodically (e.g., for WiFi scans). Third, revisiting a previously visited area yields positive, albeit smaller rewards during the site survey. This is not the case with OP. As a result, our problem is fundamentally more challenging than OP. 

In this paper, we formulate and show the NP-hardness of IPP for location fingerprint collection. We devise a greedy algorithm and a genetic algorithm (GA). The greedy algorithm picks the next waypoint (vertex) to travel to based on the ratio of marginal reward and marginal cost. In the GA we customize chromosomes, selection and mutation operations to solve IPP. Additionally, we also extend two state-of-the-art algorithms to OP to handle edge rewards and revisits.

 Due to signal variations in indoor environments, comparing the utility of data collected from different paths is challenging. Even collecting fingerprints along the same path at different times can provide different data. We propose a novel method that combines both real-world data collected in two indoor areas, and synthetically generated data to evaluate the utility of the proposed algorithms. Experimental results show that Greedy has the shortest run time, but it suffers from poor performance in MI or localization accuracy. In contrast, GA has good performance consistently in all scenarios though at the expense of higher computation complexity than Greedy. The baseline algorithms by extending known algorithms for OP suffer from either high computation complexity for large areas, sub-optimal performance, or both.

The rest of this paper is organized as follows. In Section~\ref{sect:relatedwork} we review the state of the art approaches to OP. In Section~\ref{sect:formalization} we first provide background on GPs and MI, and then define the IPP problem rigorously. The two heuristic algorithms are presented in Section~\ref{sect:algo}. Experimental results from two indoor environments are presented in Section~\ref{sect:exp} followed by discussion and conclusion in Section~\ref{sect:discussion} and Section~\ref{sect:conclusion}.

\section{Related Work}
\label{sect:relatedwork}
IPP has gained interest lately with the availability of affordable robotic platforms such as drones and rovers, but the problem is largely unsolved. A popular approach to IPP is to discretize the search space and formulate it as an OP~\cite{vansteenwegen2011orienteering}. Formally, given a graph, OP aims to determine a subset of nodes to visit, and the order to visit, so that the total collected nodal reward is maximised within a given budget. There exisits many solution to OP including: randomized algorithms ~\cite{binney2010informative}\cite{hollinger2014sampling}\cite{singh2007efficient}, branch and bound~\cite{binney2012branch}, a greedy algorithm incrementally only using a portion of the budget~\cite{hollinger2009efficient}. Under certain assumptions and relaxations, mixed integer OP can be formulated as an integer programming problem and solved accordingly~\cite{vansteenwegen2011orienteering}. Genetic algorithms have also been investigated~\cite{tasgetiren2002genetic}. In genetic algorithms (GA) each path is encoded as a chromosome. Multiple paths are then created by mutating the already existing chromosomes. Many variants of OP have been investigated in literature. For instance when multiple agents are available to collect data, it is called IPP with multiple robots~\cite{singh2007efficient}\cite{yu2016correlated}. Another approach to solve the IPP problem is to avoid discretization altogether, and instead allow a continuous path, along the trajectory that most reduces variance, similar to gradient descent~\cite{singh2010modeling}, or by using  Bayesian Optimization~\cite{marchant2014bayesian}.

In this paper, we consider IPP using a single robot (or user) by representing a target area with a graph. In addition to proposing two new algorithms to this problem, we also extend state-of-the-art methods of OP. Next, we discuss the two methods in more details.

\subsection{Recursive Greedy Algorithm}
The Recursive Greedy (RG) algorithm is an approximate algorithm for the submodular orienteering problem (SOP), first proposed by Chekuri \textit{et al.} in 2005~\cite{chekuri2005recursive}. The algorithm considers all possible combinations of intermediate vertices and budget, and then it is recursively applied on the smaller sub-problems. The greedy aspect of the algorithm comes from the divide-and-conquer approach by simply concatenating the paths returned from sub-problems into a single one as the final solution. The algorithm can be summarized informally in the following steps:
\begin{itemize}
    \item enumerate all possible combinations of intermediate vertices and budget splits
    \item recursively find the first half of the path within the budget split 
    \item find the second half of the path within the remaining budget 
    \item return the concatenation of the two sub-paths which have the biggest reward
\end{itemize}

The run time of RG is $O((2nB)^I \cdot T_f)$, where $n$ is the number of vertices in the graph and $T_f$ is the maximum time to evaluate the reward function on a given set of vertices, and $I$ is the recursion depth. The algorithm can be further improved by leveraging binary search to guess how the budget should be split, which is the Recursive Greedy Quasi-Polynomial (RG-QP) algorithm. However, the run time complexity of RG-QP is $O((2+nAlogB)^I \cdot T_f)$, where $A$ is the assumed upper bound of the best path utility. If the number of the vertices in the optimal path $\mathcal{P}^*$ is $k+1$, the utility of the path $\mathcal{P}$ returned by the recursive greedy algorithm satisfies $f_X(\mathcal{P}) >= f_X(\mathcal{P}^*) / \left \lceil{1+log(k)}\right \rceil$ when $ I >= \left \lceil{1+log(k)}\right \rceil$.

Even with RG-QP, the run time is long for practical problems, especially when the number of vertices or the budget is large. As presented in the experiments in~\cite{binney2010informative}, even with a small graph of 16 nodes, it takes many hours or even days when the budget is large. Furthermore, the run time is sensitive to the recursion depth $I$ since it is exponential with respect to $I$. As a result, the recursive greedy algorithm can be only used to solve the OP when the number of vertices and the budget is small.  

\subsection{Random Orienteering (RO)}
RO was originally introduced in~\cite{arora2017randomized}. The basic idea is to convert OP into two sub-problems: the subset selection problem and the travelling salesman problem  (TSP). Specifically, vertices are randomly added and removed from the current vertex set, and path planning is done via a TSP solver among the selected vertex set. Such an approach requires that the graph is fully connected. Additionally, the utility is additively associated with the vertices and costs arise from the edges. Thus, once the set of vertices is determined, the utility is also determined. It suffices to use a TSP solver to find the least cost path to traverse those vertices. 

\section{Problem Description}
\label{sect:formalization}
In this section, we first give a brief introduction GPs and MI. Then we formalize the problem and show the NP-hardness of IPP.

\subsection{Gaussian Processes}
\label{subsect:gp}
A GP is a stochastic process of random variables indexed by time or space, where any finite subset are jointly Gaussian~\cite{rasmussen2006gaussian}. Specifically, a GP is defined by a mean function  $m(\mathbf{x})$ and a covariance function $k(\mathbf{x}_p,\mathbf{x}_q)$. Given $f$ is a function drawn from this GP, for any finite subset of variables $\mathbf{x}_1,...,\mathbf{x}_n$, the respective function values $f(\mathbf{x}_1),...,f(\mathbf{x}_n)$ follow a joint multivariate Gaussian distribution,
\begin{equation*}
\mathcal{N} 
\begin{pmatrix}
\begin{bmatrix}
m(\mathbf{x}_1)\\
\vdots\\
m(\mathbf{x}_n)
\end{bmatrix}\!\!\ ,&
\begin{bmatrix}
k(\mathbf{x}_1,\mathbf{x}_1) & ... & k(\mathbf{x}_1,\mathbf{x}_n)\\
\vdots & \ddots & \vdots\\
k(\mathbf{x}_n,\mathbf{x}_1) & ... & k(\mathbf{x}_n,\mathbf{x}_n)
\end{bmatrix}
\end{pmatrix} .
\end{equation*}
The mean function $m$ can be any valid function. The covariance matrix $K$ is generated by the covariance function, and it is a positive semidefinite matrix. In this paper we utilize the exponential covariance function
\begin{equation*}
\label{eq:exp_kern}
k(\mathbf{x}_p,\mathbf{x}_q)= \sigma_f^2 exp(-\frac{||\mathbf{x}_p -\mathbf{x}_q||}{l}) ,
\end{equation*}
where $\sigma_f^2$ is the signal variance and $l$ is the length scale. The signal variance term controls the maximum covariance between variables, and the length scale term controls the smoothness.

A fundamental property of Gaussian distribution is that any conditional of a Gaussian distribution is still Gaussian. Suppose we model the fingerprints as a GP, $X$ is the set of observed locations and $\mathbf{y}$ is the corresponding fingerprint vector. Then, for any unobserved locations $X_*$ the posterior distribution of the fingerprints is $\mathcal{N}(\mathbf{\mu}_*,\Sigma_*)$, where
\begin{multline}
\label{eq:gp_mu}
\mathbf{\mu}_* =m(X_*) + K(X_*,X)(K(X,X)+\sigma_n^2 I)^{-1}\\(\mathbf{y} - m(X)),
\end{multline}
\begin{multline}
\label{eq:gp_sigma}
\Sigma_* = K(X_*,X_*) - K(X_*,X)(K(X,X) + \sigma_n^2 I )^{-1}\\K(X,X_*).
\end{multline}
Here $\sigma_n$ represents the noise of observations which also follow a Gaussian distribution.

\subsection{Mutual Information}
\label{subsect:mi}
In information theory, informativeness or uncertainty can be characterized by entropy. For a discrete random variable $X$ with distribution $p(x)$, the corresponding entropy is defined by 
\begin{equation}
H(X) = - \sum_{x} p(x) log p(x).
\end{equation}
For a continuous variable, the summation should be replaced with integral. Specifically, if the variable is a vector with $d$ dimension and follows a multivariate Gaussian distribution with $\Sigma$ as the covariance matrix, its entropy (also referred to as differential entropy  or continuous entropy) is given by 
\begin{equation}
\label{eq:gpentropy}
H(X) =\frac{1}{2} ln |\Sigma| + \frac{d}{2}(1 + ln (2\pi)).
\end{equation}

Given two random variables $X$ and $Y$, MI measures the mutual dependence between the variables, and it can be expressed as 
\begin{equation}
\label{eq:mi}
MI(X;Y) = H(X) - H(X|Y).
\end{equation}

The use of MI as an  informativeness measure has been studied in the sensor placement problem in \cite{guestrin2005near}. The intuition behind MI is that it puts a greater reward on locations that most significantly reduce uncertainty about locations where no measurements were taken.

It is interesting to notice that the entropy and MI depend only on the covariance matrix, which is determined by the covariance function. When the fingerprints are modeled with a GP with known hyperparameters, for any two location sets their MI can be calculated directly without taking measurements at those locations. This is the reason why the paths can be planned offline. However, in order to estimate the hyperparameters $\sigma_n, \sigma_f$ and $l$ of the exponential kernel \eqref{eq:exp_kern}, a small amount of pilot data~\cite{binney2010informative,guestrin2005near} is needed.

\subsection{Problem Formalization}
Given an area $A$ of interest, we first represent it as a graph $G = (V,E)$, where $V$ are vertices associated with valid unobstructed physical locations in the space of interest, and $E$ is the set of edges along which a user can travel. Representing a continuous area by a connected graph for path planning is an interesting problem in itself. However, in this work, we assume the graph is given. Unlike \cite{chekuri2012improved}, the graph is not restricted to be fully connected. Suppose $v_s,v_t \in V$ are the start and terminal vertices. A walk is described by a list of edges $\mathcal{P} = [e_1,e_2, ..., e_k]$, where $e_1$ and $e_k$ are edges starting and ending at $v_s$ and $v_t$, respectively. It can also be represented  equivalently as a list of nodes $\mathcal{P} = [v_s, v_i,...,v_j, v_t]$. Note that we allow revisits of vertices in the walk.

A function $f(\mathcal{P})$ is used to describe the informativeness of $\mathcal{P}$. In this work, mutual information is adopted as a measure of informativeness. Suppose a robot moves along a path $\mathcal{P}$ with a speed of $v$ \footnote{Here, we make the simplified assumption that turning of the robot costs no extra time.}, and the fingerprints are sampled every $t$ seconds. Then every $vt$ interval a sample can be taken starting from the vertex $v_s$. We represent the fingerprints associated with the vertex set $V$ as ${\mathbf y}_V$, and those at the sample locations taken by the robot by ${\mathbf y}_R$. As discussed in Section~\ref{subsect:mi}, the MI does not depend of the specific measurements but the locations (and kernels), then $f(\mathcal{P})$ can be calculated as 
\begin{equation}
f(\mathcal{P}) = MI({\mathbf y}_V;{\mathbf y}_R) = H({\mathbf y}_V) - H({\mathbf y}_V|{\mathbf y}_R).
\end{equation}
Here $H({\mathbf y}_V|{\mathbf y}_R)$ is the conditional entropy of ${\mathbf y}_V$ given the sampled locations $R$, which can be calculated from \eqref{eq:gp_sigma} and \eqref{eq:gpentropy}. 

Meanwhile, we also consider the informativeness of the pilot data used to estimate the GP hyperparameters. With a small abuse of notation, we denote the pilot data collected from location $Y_D$ as $D$ and define $f_D(\mathcal{P})$ as 
\begin{equation}
f_D(\mathcal{P}) = MI({\mathbf y}_V;{\mathbf y}_R \cup D) = H({\mathbf y}_V) - H({\mathbf y}_V|{\mathbf y}_R \cup D).
\end{equation}

The cost of traversing a path $\mathcal{P}$ is the sum of the lengths of the edges in the path 
\begin{equation}
C(\mathcal{P}) = \sum_{i=1}^{k} Length(e_i).
\end{equation}

The optimization goal is to find a path $\mathcal{P}^*$, 
\begin{equation}
\label{eq:maximization}
\mathcal{P}^* = \arg \max_{\mathcal{P} \in \Psi} f_D(\mathcal{P})~s.t.~ C(\mathcal{P}) \leq B,
\end{equation}
where $\Psi$ is the set of all possible paths in $G$ starting at $v_s$ and terminating at $v_t$, and $B$ is a finite budget (travel distance) which limits the length of the path. Clearly the budget $B$ can be also specified in travel time, in which case, the right hand side of \eqref{eq:maximization} becomes $B/v$.

\subsection{NP-hardness of IPP}
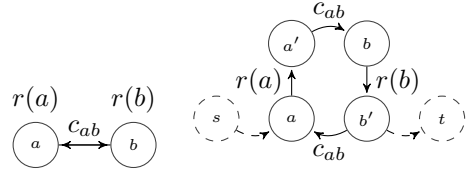
\begin{figure}[!t]
\centering
\begin{tikzpicture}[->,>=stealth',shorten >=1pt,auto,node distance=1.3cm]

\tikzstyle{every state}=[circle,draw=black!75,minimum size=6mm]
  \node[state]  (A) [font=\fontsize{6}{6}\selectfont, label=above:$r(a)$] {$a$};
  \node[state]  (B) [right of=A] [font=\fontsize{6}{6}\selectfont, label=above:$r(b)$] {$b$};

    \path (A) edge                    node {$c_{ab}$} (B)
          (B) edge                    node {} (A);
\end{tikzpicture}
\hspace{2mm}
\begin{tikzpicture}[->,>=stealth',shorten >=1pt,auto,node distance=1cm]
\tikzstyle{every state}=[circle,draw=black!75,minimum size=6mm]
    \node[state] (A)                                  [font=\fontsize{6}{6}\selectfont] {$a$};
    \node[state] (A') [above of=A]                      [font=\fontsize{6}{6}\selectfont] {$a'$};
    \node[state] (s)  [left of=A,  dashed]  [font=\fontsize{6}{6}\selectfont] {$s$};
    \node[state] (B)  [right of=A']                    [font=\fontsize{6}{6}\selectfont] {$b$};
    \node[state] (B') [below of=B]                      [font=\fontsize{6}{6}\selectfont] {$b'$};
    \node[state] (t)  [right of=B', dashed] [font=\fontsize{6}{6}\selectfont] {$t$};
    
  \path (A) edge              node {$r(a)$} (A')
        (B) edge              node {$r(b)$} (B');

  \path (s) edge    [dashed, bend right]         node {} (A)
        (A') edge   [bend left]                 node {$c_{ab}$} (B)
        (B') edge   [bend left]            node {$c_{ab}$} (A)
        (B') edge   [dashed, bend right, left]   node {} (t);
\end{tikzpicture}

 \caption{An example of the transformation from the OP graph $G$ to the IPP graph $G'$, where the starting node is $a$ and terminating node is $b$ to the equivalent $G'$. In the transformation two dummy nodes with zero reward and cost, $s$ and $t$ are added.}
 \label{fig:graphtransform}
\end{figure}

IPP can be shown to be NP-hard by reducing an instance of OP to an instance of IPP within polynomial time. The input of an OP instance is defined on a graph $G=(V,E)$, with a vertex reward function $r(v)$, and an edge cost function $c(e)$. The start and terminal vertices are $v_s$ and $v_t$, and the travel cost is limited by a budget $B$. A reduction is possible in which the optimal solution to IPP implies an optimal solution to OP as follows\footnote{We deem the undirected graph $G$ as a bidirected graph.}:
\begin{itemize}
    \item A graph $G'=(V',E')$ is constructed for IPP. Initially, $V' = V$, and then for each $v \in V$, a 'shadow' vertex $v'$ is created by making a copy of $v$.  
    \item To construct $E'$ in $G'$, for each edge $e_{ab} \in E$ in OP, we connect $(a,a'), (a',b), (b,b'), (b',a)$ as directed edges.
    \item In $G'$, we assign the reward associated with each edge $(v,v')$ as $r(v)$ in OP, and all other edges have a 0 reward. Furthermore, the cost of $(v,v')$ is defined as 0, and all other edges have the same cost as in OP.
    \item Two dummy vertices with zero edge cost and zero reward, $s$ and $t$ are added to accommodate the corresponding start and terminal vertices in OP.
\end{itemize}

Fig.~\ref{fig:graphtransform} shows an example transformation for a simple graph. In the converted graph $G'$, for every $v$ the surrounding vertices are $v'$, and for every $v'$, the surrounding vertices are $v$, which forces the solution to be a sequence of vertices alternating between $v$ and $v'$. With such a transformation, it is easy to show that the solution in $G$ for OP is optimal if and only if the corresponding solution in $G'$ for IPP is optimal.

Assume the optimal IPP solution is $[a, a', b, b', ..., k, k']$ (dummy vertices $s$ and $t$ are omitted, $a$ and $k$ are the start and terminal vertices) in $G'$, then the corresponding optimal OP solution in $G$ is $[a, b, ..., k]$. If the optimal solution for OP is not this path, but rather any other path $[a,x,...,k]$ with a larger reward, we can always construct a better path $[a, a', x, x', ..., k, k']$ in $G'$, which leads to a contradiction with $[a, a', b, b', ..., k, k']$ is the optimal IPP solution. Conversely, for any optimal OP solution in $G$, we can similarly construct the optimal IPP solution in $G'$. Since the reduction can be constructed in polynomial time and OP is NP-hard, hence IPP is also NP-hard.

\section{Informative Path Planning Algorithms}
\label{sect:algo}
From Section~\ref{sect:formalization}, we know that IPP is NP-hard. Therefore, we resort to heuristic algorithms to solve the problem. Specifically, we first present a Greedy algorithm based on the Steiner TSP solver, and then we introduce GA to solve IPP.

\subsection{Greedy Algorithm}
Greedy algorithms are known to achieve constant approximation ratios~\cite{guestrin2005near} for submodular optimization problems. The IPP problem, on the other hand, is more challenging since its budget constraints are defined on a graph.  The RG algorithm in \cite{chekuri2005recursive} is a pseudo-polynomial algorithm with logarithmic approximation ratio for submodular OPs. However, it has a long runtime in practice even for small problems.  In this section, we consider a simple greedy algorithm with low computation complexity which adds vertices incrementally. 

The first step is to define the greedy criteria. Suppose the current path planned is $\mathcal{P}$, and let $V(\mathcal{P})$ be the vertices in $\mathcal{P}$. For each candidate vertex $v_c$ that is not contained in the current path, the shortest path to traverse $V(\mathcal{P}) \cup \{v_c\}$ is denoted by $\mathcal{P}_c$. Then the marginal benefit-cost ratio (MBCR) of extending the current path to $v_c$ is given by $MBCR(\mathcal{P},v_c) = \frac{f_D(\mathcal{P}_c) - f_D(\mathcal{P})}{PathLength(\mathcal{P}_c) - 
PathLength(\mathcal{P})}$. The greedy algorithm then selects the vertex that has the highest MBCR among all remaining vertices. In computing both the utility and the cost of adding an extra vertex, the Steiner TSP (STSP) from source $v_s$ to terminal $v_t$ is solved.

The Greedy algorithm is outlined in Algorithm ~\ref{alg:ge}. The complexity is mainly determined by the TSP algorithm. For instance, if the TSP algorithm has a complexity of $O(n^2 * 2^n)$ (dynamic programming based TSP algorithm), then the complexity of the Greedy algorithm is $O(\frac{B}{e_{min}} * n^3 * 2^n)$, where $B$ is the budget and $e_{min}$ is the minimum edge length.

It is easy to construct an example where Greedy performs arbitrarily bad. However, through evaluation, we see that under certain budgets, Greedy can achieve competitive results, as shown in Fig.~\ref{fig:area2comp}.

\begin{algorithm}
\SetAlgoLined
\caption{Greedy Algorithm}
\label{alg:ge}
\SetKwInOut{Input}{Input}
\SetKwInOut{Output}{Output}
\Input{the problem Graph $G$ and the budget $B$ \\the start and end vertices $v_s$ and $v_t$ \\the pilot data set $D$ }
\Output{the vertex order of the returned path}

$\mathcal{P} = ShortestPath(v_s,v_t)$ \\

\While{$PathLength(\mathcal{P}) <= B $}
{
    \ForEach{$v_c \in V \wedge v_c \not \in V(\mathcal{P})$}{
    $\mathcal{P}_c = SteinerTSP(V(\mathcal{P}) \cup \{v_c\})$ \\
    $MBCR(\mathcal{P},v_c)  = \frac{f_D(\mathcal{P}_c) - f_D(\mathcal{P})}{PathLength(\mathcal{P}_c) - PathLength(\mathcal{P})}$
    }
    $v_{best} = \arg \max(MBCR(\mathcal{P},v_c) ) $ \\
    $\mathcal{P}_{best} = SteinerTSP(V(\mathcal{P}) \cup \{v_{best}\})$ \\
    \uIf{$PathLength(\mathcal{P}_{best}) <=B$}{
    $\mathcal{P} = \mathcal{P}_{best}$
    }
    \Else{
        break
    }
}
return $\mathcal{P}$
\end{algorithm}

\subsection{Genetic Algorithm}
\label{subsect:ga}
Genetic algorithm is a powerful and efficient evolutionary algorithm which can be utilized to solve both numerical and combinatorial optimization problems. The main idea is to imitate the process of biological natural selection. Though there are no guarantees that GA will find the global optimal solution, they are likely to be close to the global optimum~\cite{mardle1999overview}. The mutation mechanism can protect the algorithm from being stuck in local optima by diversifying the population.

In GA, a chromosome is utilized to encode a feasible solution to the specific optimization problem. A pool of chromosomes is maintained, which is also known as the population. For each chromosome, a corresponding fitness score is calculated by a fitness function. The initial population is usually randomly generated. After that, an evolutionary process which aims to simulate the biological reproduction mechanism will begin iterating. In each iteration, four steps are involved: 
\begin{itemize}
    \item {\it Selection:} Individuals (also known as parents) from the population are selected based on their fitness scores.
    \item {\it Crossover:} The selected parents will reproduce new offspring by the crossover process.
    \item {\it Mutation:} Some of the offspring are selected for mutation to increase the diversity of the population.
    \item {\it Update:} The population is updated by merging the offspring and parents.
\end{itemize}
 After a number of iterations, the individual with the highest fitness score is selected as the best solution.

A large body of literature~\cite{grefenstette1985genetic,potvin1996genetic,larranaga1999genetic} can be found using GA to solve the TSP problem. Specifically,~\cite{larranaga1999genetic} summarized the attempts to solve TSP with different crossover and mutation operators. Meanwhile, GA has also been investigated to tackle the orienteering problem~\cite{tasgetiren2000genetic,karbowska2012genetic,piwonska2010genetic}. However, in~\cite{tasgetiren2000genetic,karbowska2012genetic}, the developed GA operators require that the graph is complete. 
The case where the input is an incomplete graph is considered in~\cite{karbowska2012genetic} and vertex revisit is allowed, but only the first visit will receive reward, which is not the case for IPP.

Next we discuss the details of adapting the GA framework to solve the IPP problem. 

\subsubsection{Encoding and Fitness Function}
Since the final solution of the IPP problem is a list of ordered vertices beginning with $v_s$ and end with $v_t$, the path representation is a natural choice. Specifically, each chromosome represents a solution in the form of $\mathcal{P} = [v_s,v_i,...,v_j,v_t]$. For each pair of neighboring vertices in the ordered list, there must be an edge connecting them. The fitness function is chosen to be equal to the utility function $f_D$ as discussed in Section~\ref{sect:formalization}. As such, each fitness score represents how much the uncertainty can be reduced by sampling along the path $\mathcal{P}$, which is directly linked to the optimization objective. 

\subsubsection{Initializing Population}
In GA, the initial population is usually generated randomly to diversify the individuals. When using GA to solve TSP on a fully connected graph, the initial population can be initialized with the random permutations of all the vertices since each vertex is visited only once and the solution is a tour without start and end locations. However, when initializing the population for IPP, three constraints must be satisfied. Specifically, i) the start and end vertices are specified; ii) the budget limitation is satisfied; iii) each vertex is allowed to be visited multiple times.

\begin{algorithm}
\SetAlgoLined
\caption{  GA Population Initialization}
\label{alg:gainit}
\SetKwInOut{Input}{Input}
\SetKwInOut{Output}{Output}
\SetKwRepeat{Do}{do}{while}
\Input{the problem Graph $G$ and the budget $B$ \\the start and end vertices $v_s$ and $v_t$\\the population size $popsize$}
\Output{the initial population}

$poppool = [];$ \\
\While{$sizeof(poppool) < popsize$}
{
    $seq = [v_s]$ \\
    \Do{$length(seq) < 0.5 * B $}{
        $v_{last} = $ last vertex in $seq$ \\
        $v_{adj}=$ sample a vertex from $v_{last}$'s neighbors\\ 
        append $v_{adj}$ to seq \\
        \uIf{$length(seq + v_{adj}) > 0.5 * B $}{
            delete $v_{adj}$ from seq \\
            break
        }
    }
    $v_{mid}= $ last vertex in $seq$ \\
    $seq2 = shortestpath(v_{mid},v_t)$ \\
    \Do{$length(seq2) < 0.5 * B$}{
           $v$ = sample a vertex from $V - (seq \cup seq2)$ \\
           insert $v$ into $seq2$\\
           /*Note the insertion location is the position which will cause the minimum budget increase. If there is no direct edges between two vertices, shortest path is utilized*/\\
           \uIf{$length(seq2) >0.5 * B$}{
               delete $v$ from $seq2$ \\
               break
           }
    }
    $chromosome = seq + seq2[1:]$ \\
    add $chromosome$ to $poppool$
}
return $poppool$
\end{algorithm}

Algorithm ~\ref{alg:gainit} describes the procedure to initialize the population. Starting from the start vertex $v_s$, an adjacent vertex to the current vertex is randomly selected and appended until half of the budget is used. The end vertex $v_t$ is then connected with the shortest path. If there is still remaining budget, vertices are sampled and inserted. If the generated chromosome by such a scheme exceeds the budget limit, then the chromosome is dropped and the procedure restarts (or adjust the budget on the first half if necessary).

\subsubsection{Selection and Crossover}
Selection simulates the idea of ``Survival of the fittest''. Typical selection methods include Roulette Wheel Selection, Rank Selection, Tournament Selection~\cite{sivaraj2011review}.  We adopt the popular tournament scheme to select parents. During each tournament, $k$ individuals are randomly selected and the winner (the one with the highest fitness score) is picked as one parent.

Crossover takes place between two selected parents. A single point crossover similar to~\cite{karbowska2012genetic,piwonska2010genetic} is utilized to generate the offspring. Specifically, common vertices (except for the start and end vertices) in the two parents are searched and one common vertex is randomly picked as the crossover point. Segments are then exchanged around the common vertex. For instance, suppose two parents $[v_s, v_1, v_2, v_5, v_7,v_8, v_t]$ and $[v_s, v_3, v_4, v_5, v_6,v_9, v_t]$ are selected, and $v_5$ is the common vertex. After crossover, two offspring $[v_s, v_3,v_4,v_5,v_7,v_8,v_t]$ and $[v_s,v_1,v_2,v_5,v_6,v_9,v_t]$ can be created. Meanwhile, only the offspring that do not exceed the budget limit can survive. If there are no common vertices then no offspring is generated. 

\subsubsection{Mutation}
\begin{figure}[!t]
	\centering
	\subfloat[\label{fig:befmutation}][before mutation]{{\includegraphics[width=0.22\textwidth]{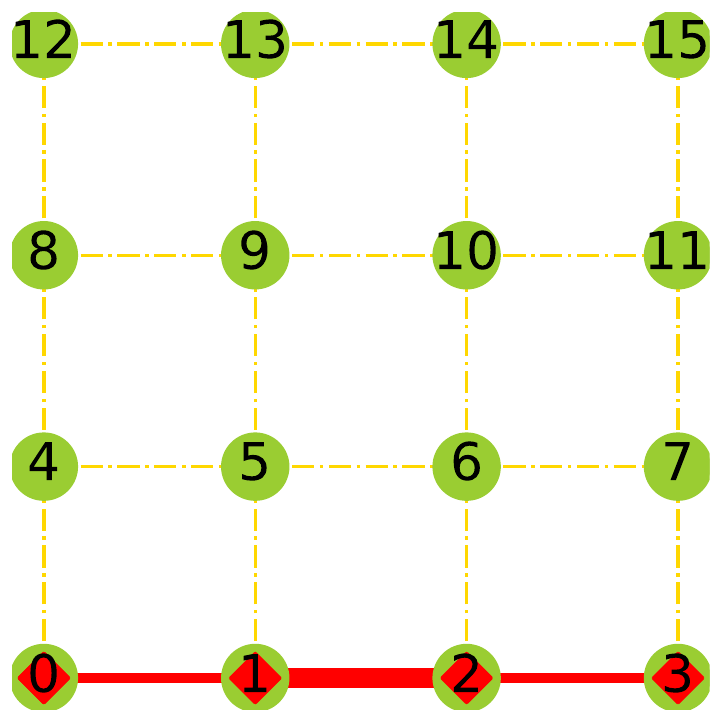} }}%
	\subfloat[\label{fig:aftmutation}][after mutation]{{\includegraphics[width=0.22\textwidth]{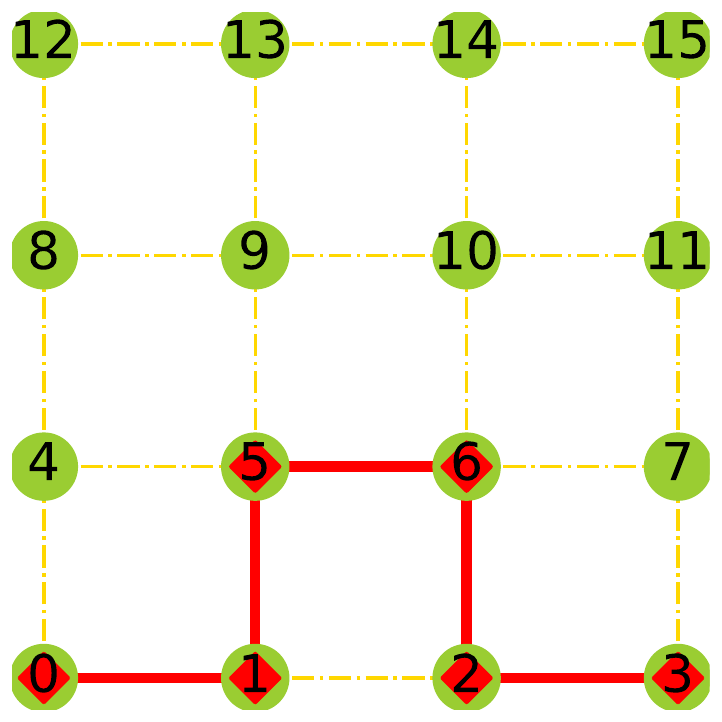} }}%
	\caption{An example of mutation. (a) shows a path [0,1,2,3] and vertices 1 and 2 are the selected mutation positions. The two vertices do not have a common adjacent vertex. However, the connection can be built through 1's adjacent vertex 5 and 2's adjacent vertex 6 as shown in (b).}
	\label{fig:mutation}
\end{figure}

Mutation mechanism is designed to promote the diversity of the population. From the optimization perspective, it enables the algorithm the ability to escape from a local optima. Various mutation operators (insertion, exchange, displacement, inversion) are developed for a TSP problem with GA when the graph is complete. When a graph is not fully connected, most of these operators is not feasible. Therefore, we propose a local extension mutation operator. Specifically, for a chromosome (a path) $\mathcal{P}$, we randomly select two intermediate adjacent vertices $v_i,v_j$ as the mutation location. The simplest case is when $v_i$ and $v_j$ have a common adjacent vertex $v_k$. Then $v_k$ can be directly inserted between $v_i$ and $v_j$. When $v_i$ and $v_j$ do not have any common adjacent vertex, the adjacent vertex set is searched to build the connection for the two segments. Fig.~\ref{fig:mutation} illustrates an example for this case.

\section{Performance Evaluation}
\label{sect:exp}
Experiments are conducted to compare the performance of different IPP algorithms and to validate the relationship between utility and localization errors. Wi-Fi RSS is collected as location dependent fingerprints and used for training localization models. The approach can be easily extended to include other types of fingerprints. Instead of a human collecting the fingerprints, a robot is utilized. 

Next, we introduce the implementation details. Then we describe the evaluation methodology and present the experimental results.

\subsection{Implementation}

All the algorithms are implemented with Python 2.7 and are run on a MacBook with Intel Core i7 with 16GB RAM\footnote{We will also use brute force approach to search for the optimal path, which is run on Compute Canada due to the extremely long run time.}.  We use GPy~\cite{gpy2014} to train and optimize the models. Graphs are created and represented by NetworkX~\cite{hagberg2008exploring}. 

In addition to Greedy and GA, we have extended the RG algorithm in~\cite{chekuri2005recursive} and RO in~\cite{arora2017randomized} to solve IPP. The detail is as follows. 

\paragraph*{Recursive Greedy Algorithm} 
We implement QP-RG as outlined in~\cite{chekuri2005recursive}. The main modification is that when evaluating $f_D(\mathcal{P})$, we take samples along the edges instead of only considering the vertices. The recursion depth parameter $I$ is set to two. We tried to set this parameter to three, but then the algorithm did not terminate for hours since the complexity is exponential with respect to $I$. Similar run time can be found in~\cite{binney2010informative} even with a graph of 16 vertices.

\begin{figure}[!t]
	\centering
	\subfloat[\label{fig:g16grid}][the original graph]{{\includegraphics[width=0.22\textwidth]{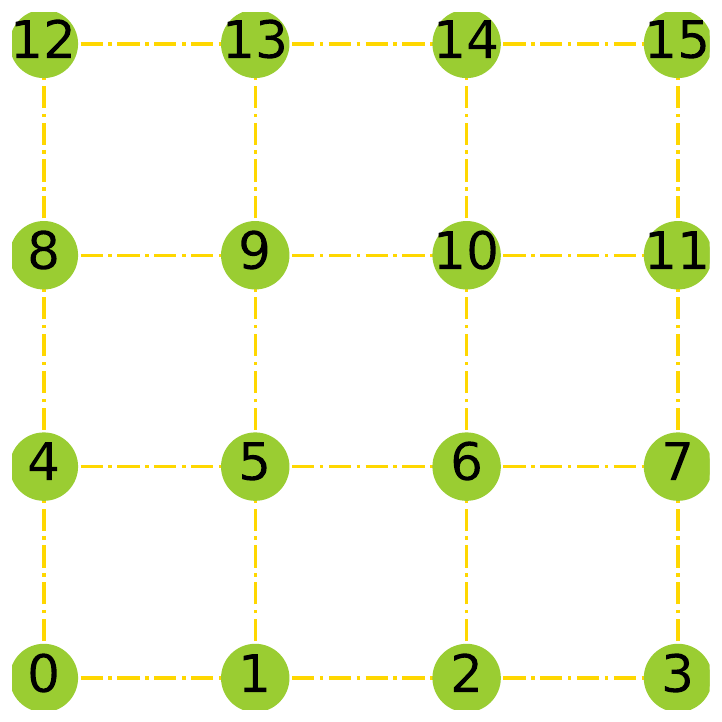} }}%
	\subfloat[\label{fig:g16edgegraph}][represent edges as nodes]{{\includegraphics[width=0.22\textwidth]{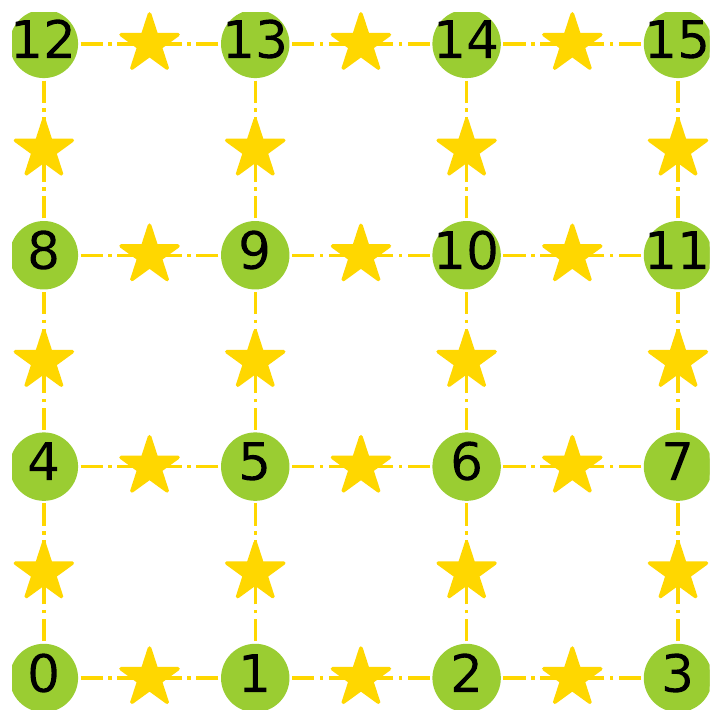} }}%
	\caption{Illustration of graph transformation for ERO. The original graph is a 4 by 4 grid graph. The gold stars represent the edge nodes.}
	\label{fig:exedgenode}
\end{figure}

\paragraph *{Edge based Random Orienteering (ERO)} 
We extended RO in~\cite{arora2017randomized} to handle edge-based rewards. The edge nodes are added or deleted randomly. Fig.~\ref{fig:exedgenode} gives an example of such a transformation. Furthermore, a Steiner TSP solver is used to plan a path among the selected edge nodes. The Steiner TSP solver is necessary since the resulting graph is incomplete. Our Steiner TSP solver is implemented based on the Concorde TSP solver~\cite{applegate2006concorde} by calculating the shortest path between every pair of vertices~\cite{letchford2013compact}. 

\subsection{Evaluation methodology}

\subsubsection{Fingerprint Collection}
\begin{figure*}[!t]
	\centering
	\subfloat[\label{fig:area1}][Area One]{{\includegraphics[width=0.3\textwidth]{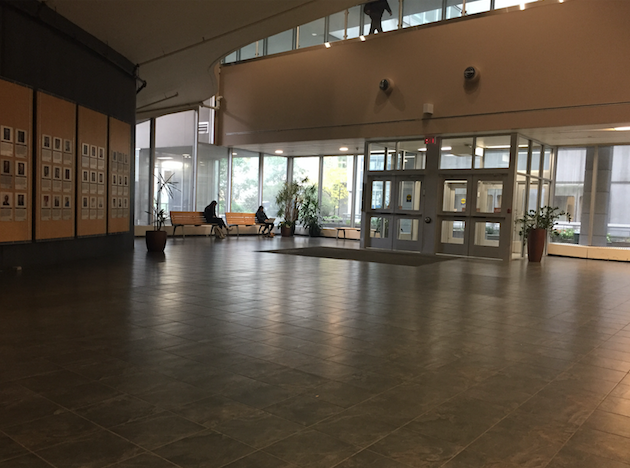} }}%
	\qquad
	\subfloat[\label{fig:area2}][Area Two]{{\includegraphics[width=0.3\textwidth]{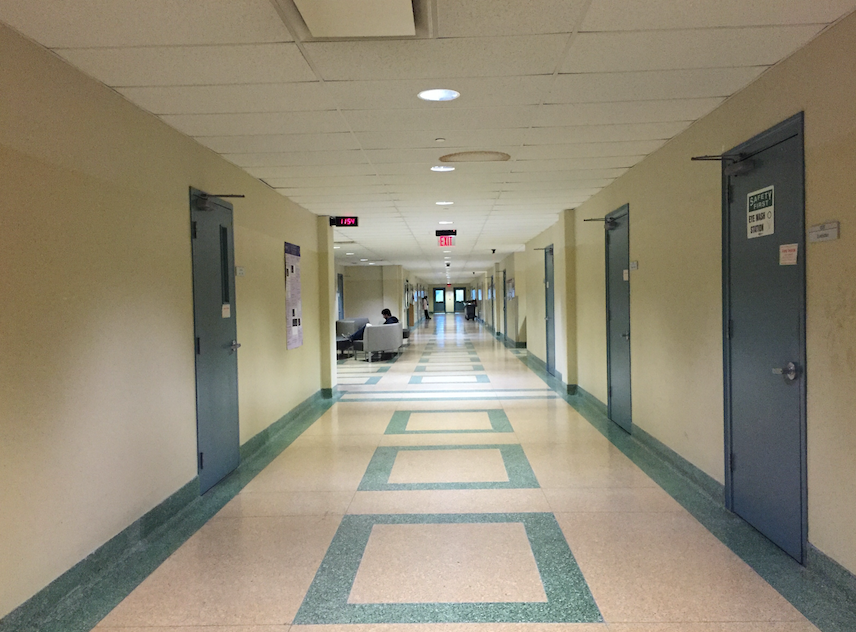} }}%
	\qquad
	\subfloat[\label{fig:robot}][The Robot]{{\includegraphics[width=0.28\textwidth]{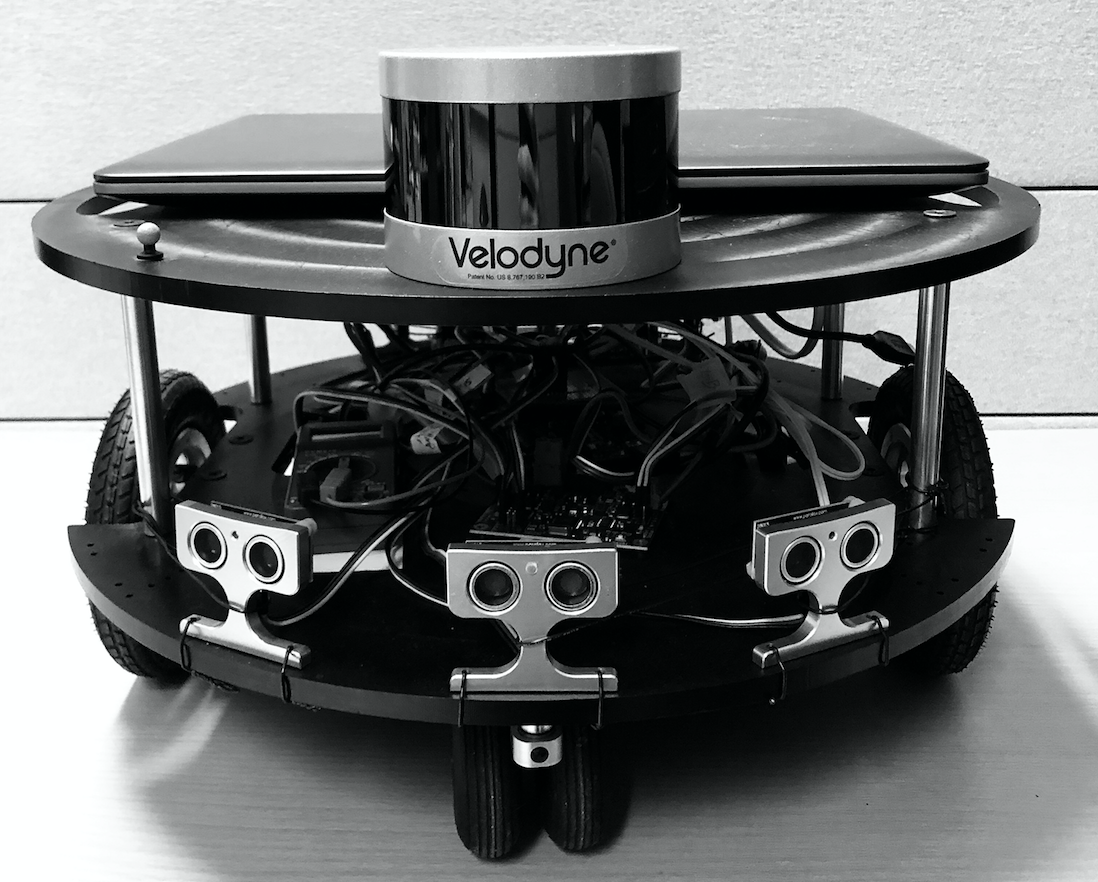} }}%
	\caption{Test areas and the robot for fingerprint collection.}
	\label{fig:areasandrobot}
\end{figure*}

\begin{figure}[!t]
	\centering
	\includegraphics[width=0.45\textwidth]{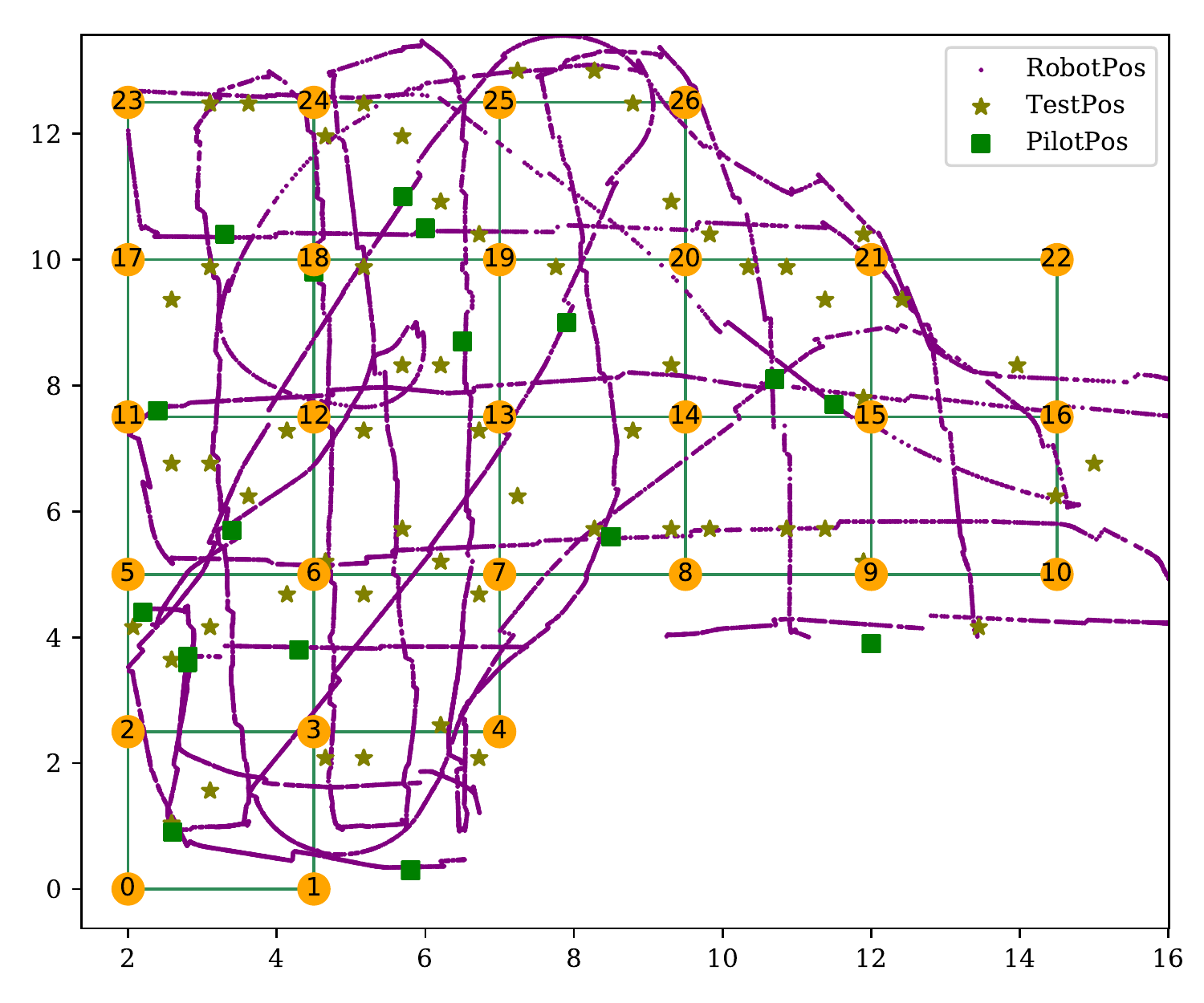}
	\caption{The graph generated from Area One. This area is discretized and represented as a grid graph. The purple lines show the robot's trajectories. The squares represent the pilot data locations, and the stars represent the test locations.}
	\label{fig:area1graph}
\end{figure}

\begin{figure*}[!t]
	\centering
	\includegraphics[width=0.95\textwidth]{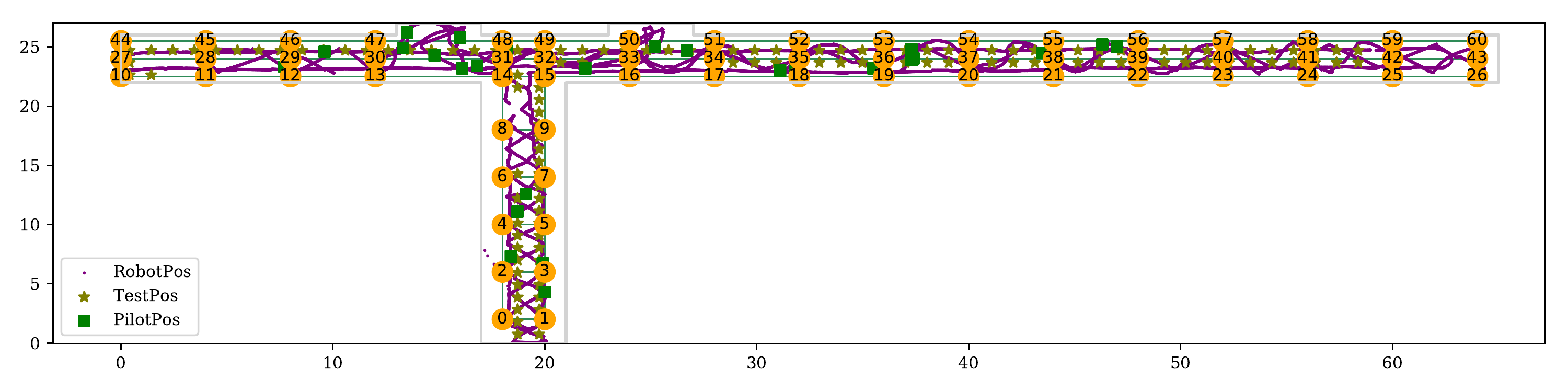}
	\caption{The graph generated from Area Two. Similar to Fig.~\ref{fig:area1graph}, the squares represent the assumed pilot data locations, the stars represent the locations for localization error evaluation and the purple lines show the rover's trajectories.}
	\label{fig:area2graph}
\end{figure*}

Two areas are selected for fingerprint collection and experiments in two buildings. The first is approximately 12m wide and 15m long. The second is a corridor and it is 63m long. Fig.~\ref{fig:areasandrobot} shows the two areas and the robot for fingerprint collection. Fig.~\ref{fig:area1graph} and Fig.~\ref{fig:area2graph} show the graphs generated from the two areas, respectively. In evaluating localization errors, test locations are selected roughly uniformly across the test areas, as shown by the stars in the figures. Table~\ref{tab:testareadetails} summarized the settings of the data collection.


\begin{table}[!t]
\caption{Information of collected data in the two areas}
\label{tab:testareadetails}
\begin{tabular}{|l|l|l|}
\hline
                     & \textbf{Area One} & \textbf{Area Two} \\ \hline
\# of APs observed           & 11                & 18                \\ \hline
\# of RSS collected     & 29115             & 199590             \\ \hline
\# of Vertices in the graph     & 27                & 61                \\ \hline
\# of Pilot Locations  & 20                 & 30                \\ \hline
\# of Test Locations & 58                & 131               \\ \hline
Localization Error (m, all RSS)  & 3.8                & 2.06               \\ \hline
\end{tabular}
\end{table}

A Wi-Fi interface card is equipped on a robot to collect the raw fingerprints. The robot is a moving platform equipped with a Velodyne Lidar allowing the robot to perform Simultaneous Localization And Mapping (SLAM), and it is manually driven to cover the available area. Simultaneously the Wi-Fi RSS measurements are recorded and labeled with the locations where they were collected, as determined by the robot's SLAM algorithm. Location errors of the robot from the SLAM algorithm, is around 15cm, depending on the number of distinct features present in the test areas. 

Compared with collecting data by human holding a smart-phone, a robot has three advantages. Firstly a human body can block the WLAN radio signal and cause a significant decay~\cite{della2012human}. Secondly the moving speed of the robot can be controlled with more precision. Lastly, the location labeling error is quite small (15cm) with the help of SLAM. However, when the data is collected by a user holding a smart-phone, the location labels are either estimated manually or generated by leveraging a step counter~\cite{li2017turf}, which are less accurate.

\subsubsection{Experimental Design}
To evaluate different path planning strategies fairly, it is important to subject them to comparable data. However, due to the time varying nature of WiFi signals, even collecting RSS along the same trajectory multiple times would result in different data. Another consideration is that during actual data collection using the robot, its speed may vary due to the presence of people in the area. To mitigate the two issues, we propose an efficient method to evaluate utility and localization errors of models trained from the fingerprints collected along different paths.

\begin{figure}[!t]
\centering
\includegraphics[width=0.47\textwidth]{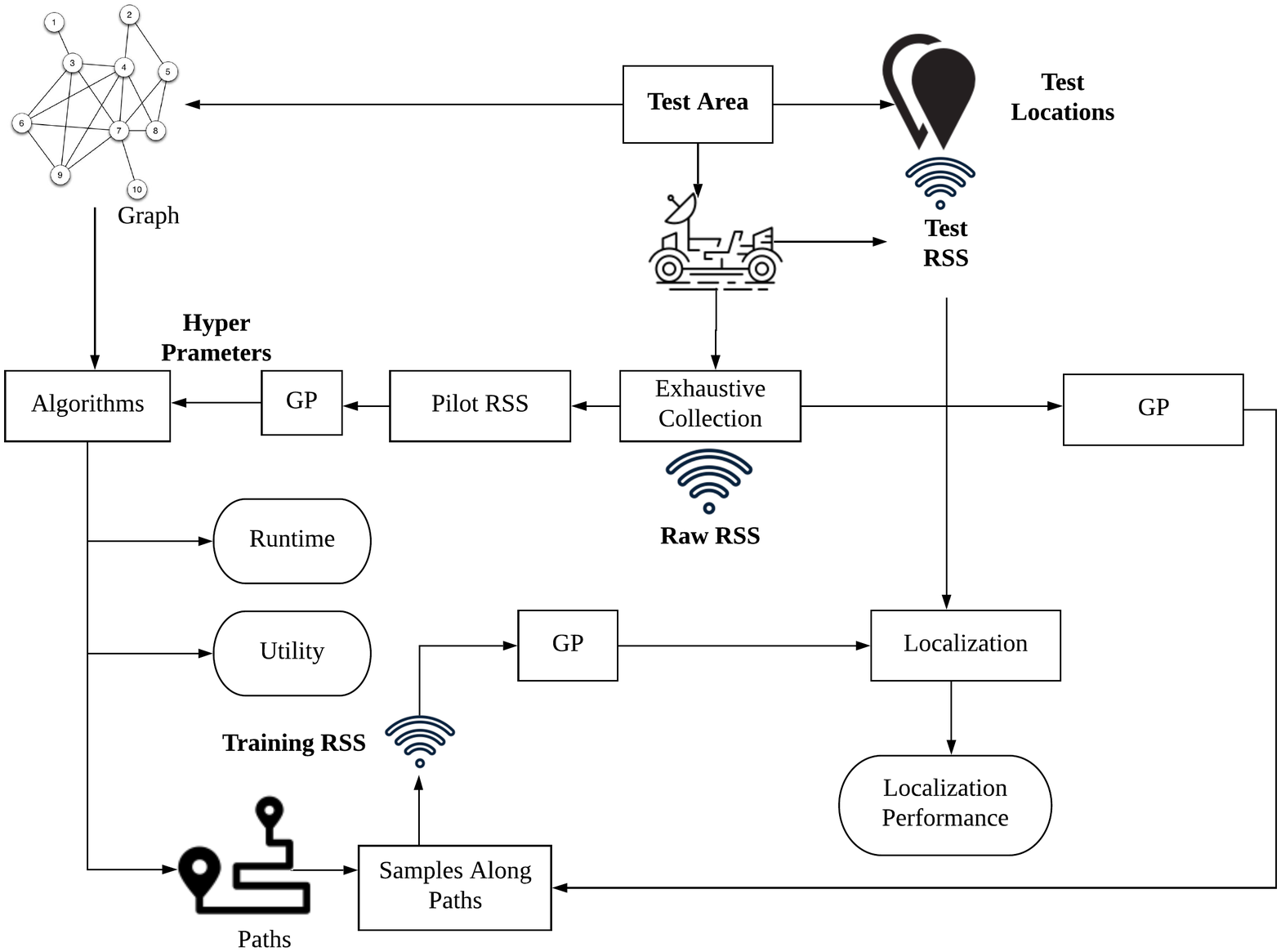}
\caption{Experimental Design}
\label{fig:expdesign}
\end{figure}

Fig.~\ref{fig:expdesign} illustrates the steps involved in the experimental design. Specifically, we exhaustively sample the test areas with the robot to collect the raw fingerprint data. Then, the raw data is used to fit a GP regression model for each observed Wi-Fi AP. The set of GP models is denoted by $\mathcal{M}$. Pilot RSS measurements are taken from this round of data collection and are utilized to learn the hyper-parameters of GP, which will be used to calculate the utility of paths. The test areas are discretized into graphs and paths are generated by different algorithms, and RSS samples are generated by the fitted GP models from the raw fingerprints. In the localization error evaluation stage, RSS measurements are collected at the test locations.

During experiments, we let $v_s$ equal to $v_t$, since we want the robot to return to the start point after finishing the fingerprint collection. Furthermore, when samples are generated along the paths, we set the sample interval to 0.5m. Sample more frequently is feasible, but it will lead to a longer time to evaluate $f_D(\mathcal{P})$, which is a subroutine to calculate the utility. Since our goal is to compare different algorithms, it is acceptable as long as all the algorithms have the same configuration.

By such a design, during the localization error evaluation stage, since the fingerprints are generated by the same models, the only factor that may affect the collected fingerprints is the specific path. Thus the environmental problems are eliminated.

\subsubsection{Performance Metric}
We consider the utility obtained, run time and the localization error by the fingerprints sampled along the paths. Utility and run time are directly returned from the path planning algorithms. 

To evaluate the localization error, we adopt the approach as in~\cite{li2017turf}. Specifically, after paths are planned, fingerprints are sampled from the GP model $\mathcal{M}$ along the paths and utilized as training data. The test area is discretized into a set of uniformly distributed reference locations. For each AP, a Gaussian Process regression model is fit based on the training data and used to predict the fingerprint distributions at those reference locations. In the inference stage, given the fingerprints collected at a test location, the target location is predicted to be one of the reference locations that has the maximum probability to generate the fingerprints. 

\subsection{Results}

\subsubsection{Choice of GA parameters}
\begin{figure}[!t]
\centering
\includegraphics[width=0.45\textwidth]{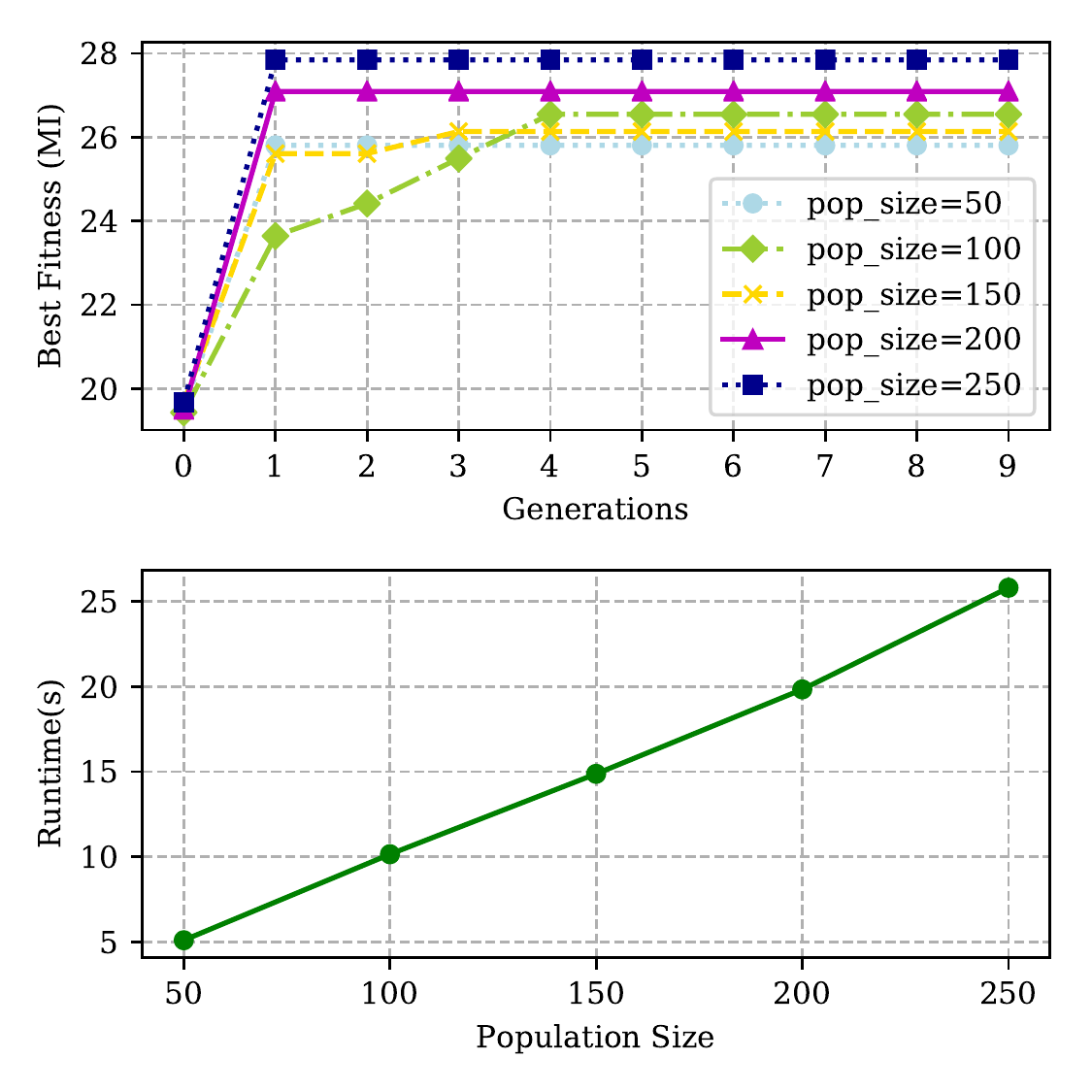}
\caption{The utility of the GA with different population size and the corresponding run time. In this experiment, tournament size is set to ten and 90\% offspring are mutated.}
\label{fig:gaperf}
\end{figure}

GA is a flexible algorithm and there are some parameters that need to be configured, including the population size $pop\_size$, the tournament size $tn$, the percentage of offspring that needs mutation $mu\_percent$ and the number of generations to iterate $g\_num$. Here, we focus on the population size and the number of generations since the run time of GA is mainly determined by these two parameters. We run GA with different $pop\_size$ on the graph derived from Area One for ten generations. Fig.~\ref{fig:gaperf} shows the best fitness during each generation and the run time of different population sizes.  As can be seen from the figure, a larger population size is more likely to achieve a higher utility, especially during the early generations. On the other hand, the run time increases as the population size becomes larger. We find that a population size of 200 gives a good trade off between run time and fitness, the parameters of GA for all experiments are listed in Table~\ref{tab:gaparam}.

\begin{table}[!t]
\centering
\caption{GA parameter setting}
\label{tab:gaparam}
\begin{tabular}{|l|l|}
\hline
 \textbf{Parameter}                    &  \textbf{Value}      \\ \hline
population size               & 100 and 200 \\ \hline
tournament size               & 10          \\ \hline
offspring mutation percentage & 90\%        \\ \hline
total generations             & 5           \\ \hline
\end{tabular}
\end{table}

\subsubsection{Relation between MI and Localization Error}
In the problem formulation in Section~\ref{sect:formalization}, we use MI as a surrogate utility measure for fingerprint data collection. The intuition is data collected along paths with higher utility will lead to models with lower localization errors. To validate this intuition, we conduct experiments to investigate the relation between utility and localization error. Specifically, we randomly sample a number of paths with different fitness scores from the GA's population of different generations, and then sample fingerprints along the paths from model $\mathcal{M}$. Localization models are trained and localization errors are evaluated on the test locations.

\begin{figure}[!t]
\centering
\includegraphics[width=0.45\textwidth]{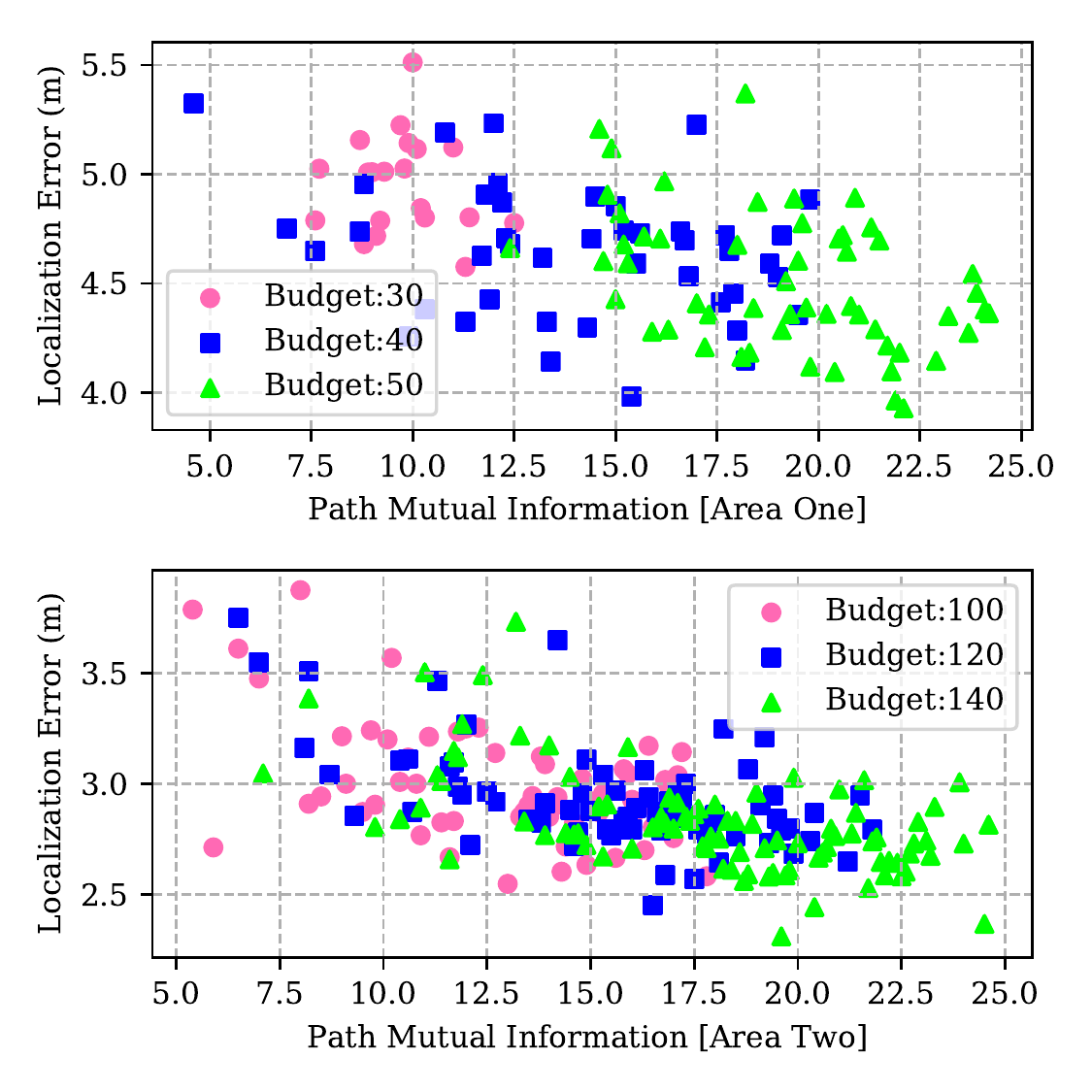}
\caption{Relation between utility and localization error in the two areas. In Area One, the Pearson correlation coefficients for budget 30, 40 and 50 are -0.12, -0.28 and -0.43, respectively. In Area Two, the Pearson correlation coefficients for budget 100, 120 and 140 are -0.51, -0.56 and -0.56, respectively.}
\label{fig:utilityvslocerr}
\end{figure}

Fig.~\ref{fig:utilityvslocerr} illustrates the utility and the corresponding localization error at the two areas under different budgets. It can be seen that the localization performance in Area One tends to be better than in Area Two. In both areas we observe that when the utility increases, the localization error tends to decrease, although it is not always true. At higher utilities (with longer paths), the localization errors become saturated. This is expected due to the high variance of fingerprints and can be analyzed through Cramer-Rao bounds~\cite{catovic2004cramer}. Intuitively, for two locations to be distinguishable, the distributions of fingerprints should have distinctive means and little overlap. Consider an extreme example, where the Wi-Fi RSS is the same everywhere. The location errors are unbounded despite more samples always increasing the MI. However, in practical scenarios as illustrated in Fig.~\ref{fig:utilityvslocerr}, the MI criteria is an effective surrogate for planning paths to collect the fingerprints.

\subsubsection{Performance}
\begin{figure*}[!t]
	\centering
	\subfloat[\label{fig:area1utility}][Utility]{{\includegraphics[width=0.3\textwidth]{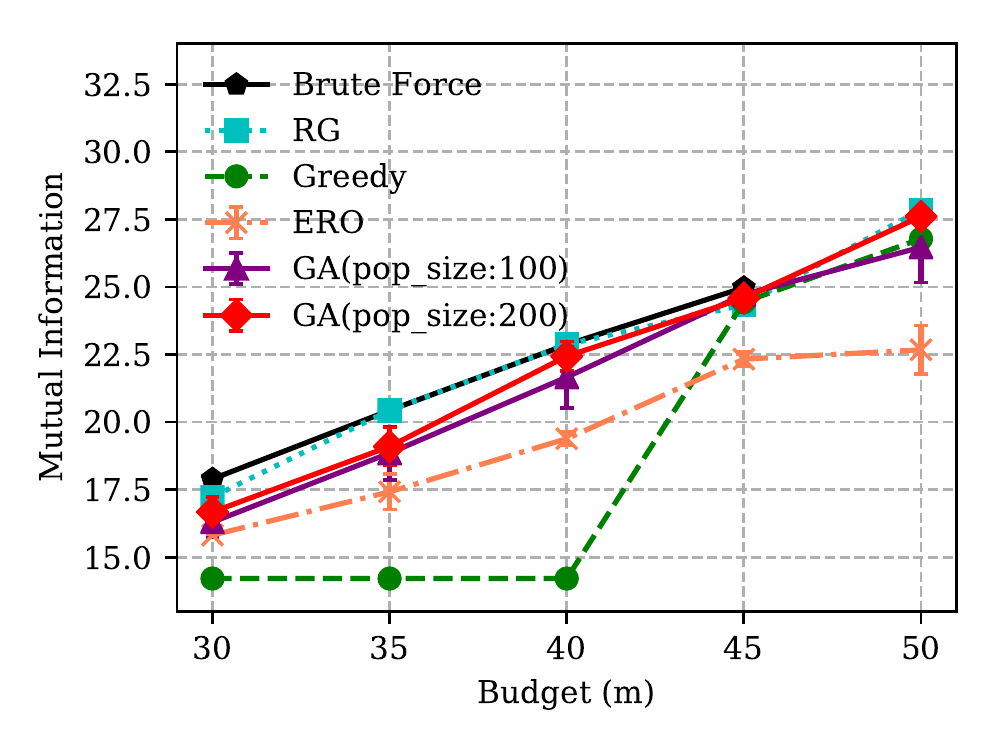} }}%
	\qquad
	\subfloat[\label{fig:area1runtime}][Runtime]{{\includegraphics[width=0.3\textwidth]{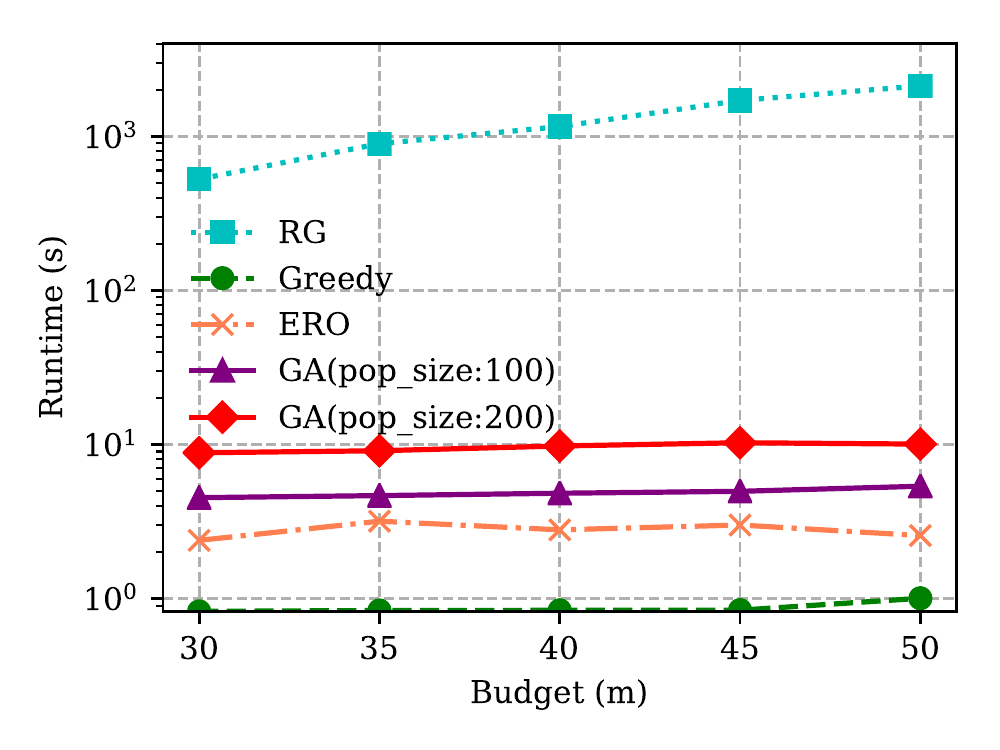} }}%
	\qquad
	\subfloat[\label{fig:area1pathlocerr}][Localization Error]{{\includegraphics[width=0.3\textwidth]{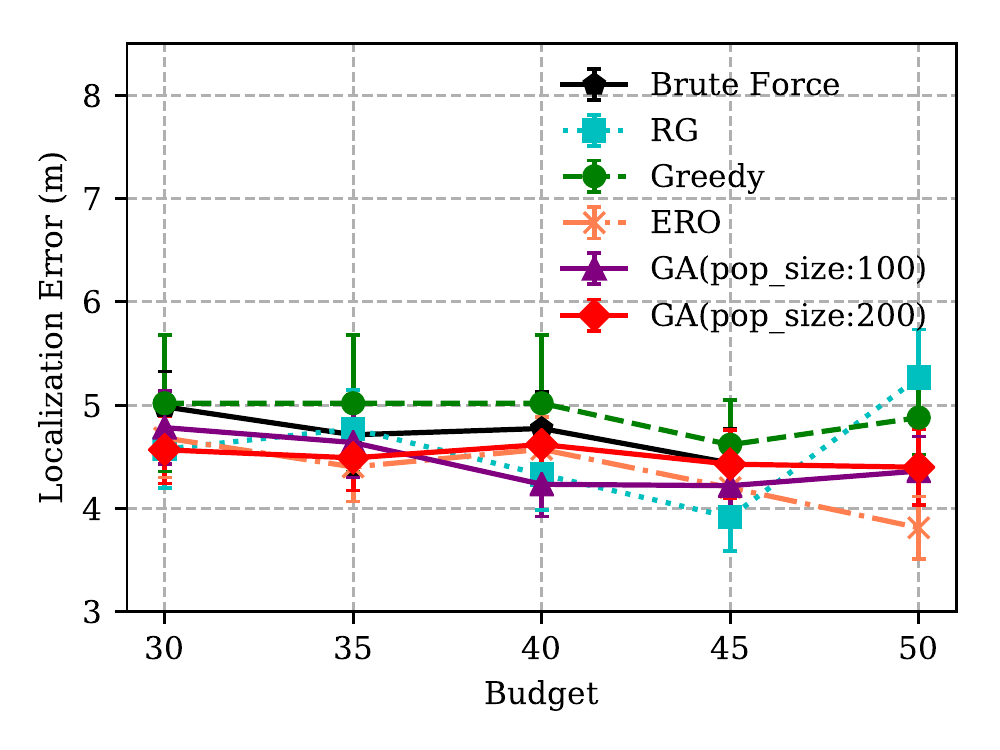} }}%
	\caption{Comparison of different algorithms in Area One under different budget constraints. The brute force approach failed to give the result in 72 hours when the budget is 50.}
	\label{fig:area1comp}
\end{figure*}

\begin{figure*}[!t]
	\centering
	\subfloat[\label{fig:area2utility}][Utility]{{\includegraphics[width=0.3\textwidth]{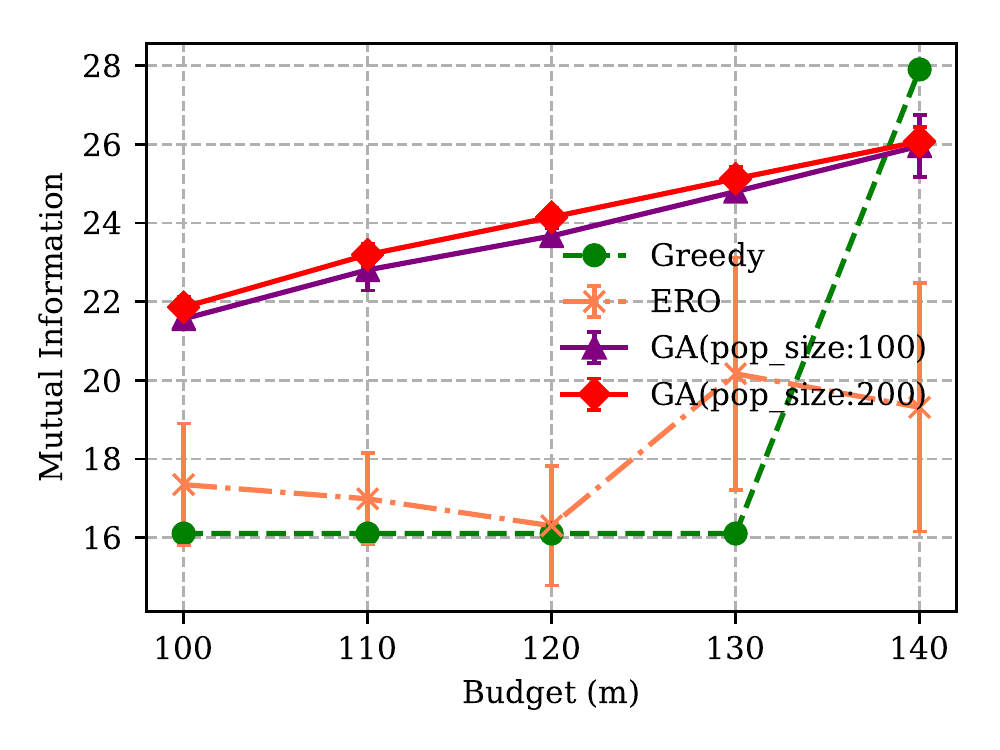} }}%
	\qquad
	\subfloat[\label{fig:area2runtime}][Runtime]{{\includegraphics[width=0.3\textwidth]{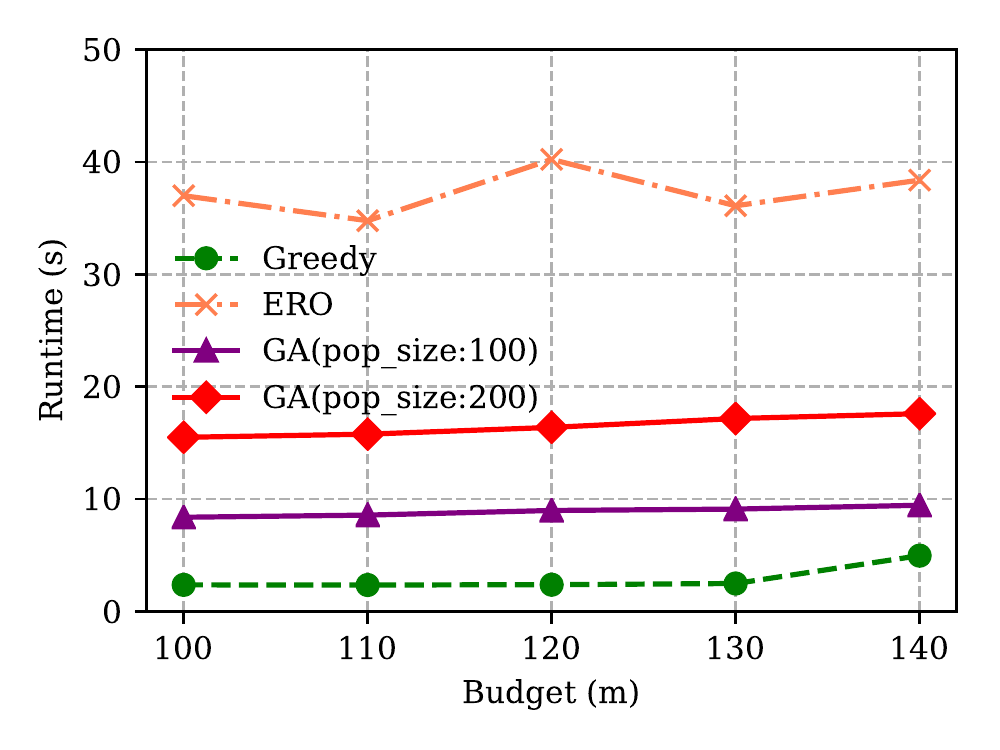} }}%
	\qquad
	\subfloat[\label{fig:area2pathlocerr}][Localization Error]{{\includegraphics[width=0.3\textwidth]{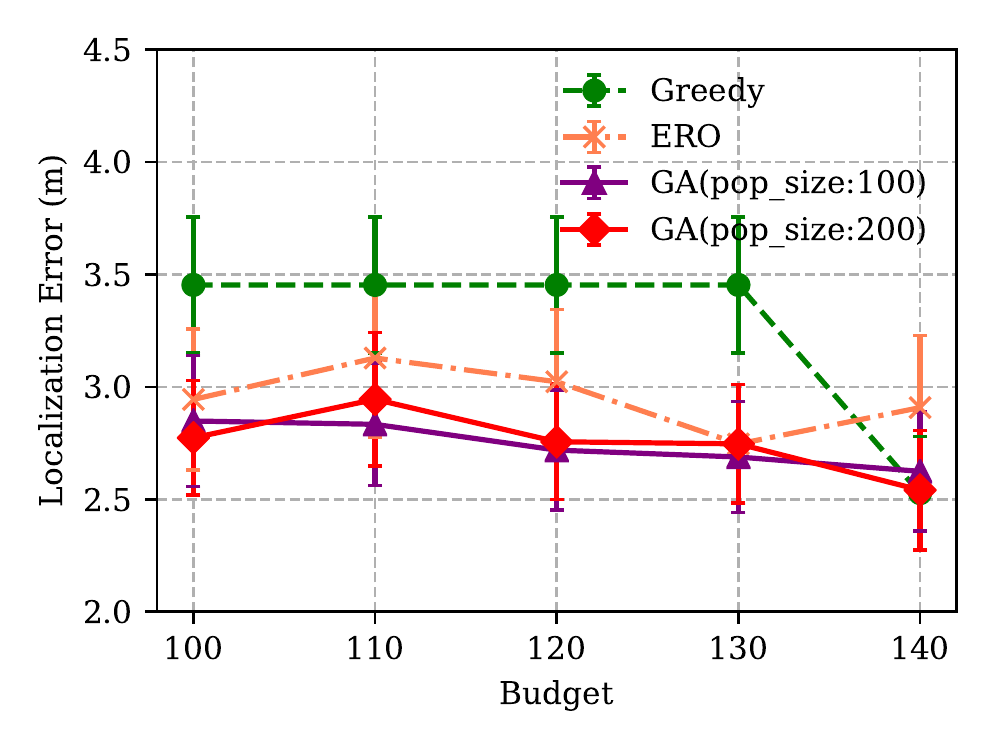} }}%
	\caption{Comparison of different algorithms in Area Two under different budget constraints.}
	\label{fig:area2comp}
\end{figure*}

 We only manage to run QP-RG in Area One, while in Area Two the algorithm failed to return after two hours due to a larger graph size and budget, thus we did not expect to run RG on large graphs. For ERO and GA, since they are randomized algorithms and each run may give different results, we run five rounds under each budget and take the average result. Fig.~\ref{fig:area1comp} and Fig.~\ref{fig:area2comp} show the results at Area One and Area Two, respectively.
 
 In order to find the path with optimal utility, we also implemented a brute force approach by enumerating all the paths constrained by the start vertex, terminal vertex and budget. The brute force approach is run on Compute Canada, and the maximum run time allowed is 72 hours for each run. In Area One, the brute force approach successfully finds the optimal path within 72 hours when the budget is set to 30,35,40 and 45.
 
\paragraph*{Utility} 
Fig.~\ref{fig:area1comp}(a) and Fig.~\ref{fig:area2comp}(a) give the utility achieved by different path planning algorithms in the two areas. We first notice that Greedy sometimes can achieve a good utility, while sometimes cannot. For example, in Area One, the utility obtained by Greedy stays the same when the budget increases from 30 to 40, and similar patterns can be found in Area Two. In contrast, the utilities attained by other algorithms grow monotonically with budgets in area One. While in Area Two, ERO tends to fluctuate. Since ERO is extended from RO~\cite{arora2017randomized}, it is not stable especially when the budget is not sufficient. Most candidate solutions are likely to exceed the budget constraints when nodes are randomly added or deleted. GA shows a promising performance in both of the two areas, and in Area One its utility is quite close to RG and the brute force approach.

\paragraph*{Runtime} 
The run time of RG increases from 527.3 seconds to 2124.7 seconds when the budget increases from 30 to 50 in Area One. The run time is quite sensitive to the budget in RG since it exhaustively search the middle vertices and budget splits. In Area One, Greedy is quite fast with less than 2 seconds, and ERO shows a faster speed than GA. However, since both Greedy and ERO invoke a Steiner TSP solver, when the size of the graph increases the run time will increase significantly. As it can be seen from~Fig.\ref{fig:area2comp} the run time of ERO increases to 40 seconds in Area Two since the corresponding graph has 61 vertices (27 vertices in Area One).  The run time of GA is mainly determined by the population size and the number of generations as shown in Fig.~\ref{fig:gaperf}. The run time of the brute force approach on Compute Canada is listed in Table~\ref{tab:bf}. When the budget rises, the run time increases dramatically, and thus for the budget of 50 it failed to give the result.
\begin{table}[!t]
\centering
\caption{Run time of brute force for Area One}
\label{tab:bf}
\begin{tabular}{|l|l|}
\hline
\textbf{Budget} & \textbf{Runtime (s)} \\ \hline
30              & 111.1                \\ \hline
35              & 1365.7               \\ \hline
40              & 37708.5              \\ \hline
45              & 133456.5             \\ \hline
\end{tabular}
\end{table}

\paragraph*{Localization Error} 
In general, the localization error decreases when the budget increases, although there are exceptions.  In most cases we see GA has a small localization error due to a large utility, since the two are correlated as shown in Fig.~\ref{fig:utilityvslocerr}.

\section{Discussion}
\label{sect:discussion}
IPP is a NP-hard problem, and there are no benchmark instances which have known optimal solutions. It is challenging to find the optimal solutions especially on a large graph with sufficient budget, since the potential solution space is extremely large.  We show that GA is flexible and well suited for solving IPP. GA can be configured to adapt to different problem scales. Increasing population size and the number of generations promises a solution with more utility, but takes a longer time. While the run time of Greedy and ERO increases significantly when the graph size increases due to the TSP solvers.

Another advantage of GA is that it can be easily parallelized. One main time consuming operation of IPP is to evaluate $f_D(\mathcal{P})$. When evaluating $f_D(\mathcal{P})$, to calculate the conditional entropy of the vertex locations based on the sample locations, involves a matrix inversion operation. GA maintains a population and for each individual $f_D(\mathcal{P})$ needs to be evaluated, and this task can be parallelized with multi-threading. In our experiments, we did not use parallelization to improve the speed.

\section{Conclusion}
\label{sect:conclusion}
In this paper, we formulated the IPP problem to plan paths for location dependent fingerprint collection. We proposed a Greedy algorithm and a Genetic Algorithm to plan paths in a graph.  Experiments were conducted to compare the performance of these algorithms. Localization experiment showed that paths with a higher utility are more likely to achieve lower localization errors, which demonstrated that mutual information is an effective surrogate for planning paths offline. In practice, we recommend to use GA to plan paths. Our future work includes planning paths online and seeking better surrogates for IPP in order to decrease the localization error.

\bibliographystyle{IEEEtran}
\bibliography{ref}
\end{document}